\documentclass[journal]{IEEEtran}

\newcommand{\matr}[1]{\mathbf{#1}}

\usepackage{amsmath}
\usepackage{graphicx}
\usepackage{subcaption}
\usepackage{xcolor,colortbl}
\usepackage{multirow}

\newcommand{\mc}[2]{\multicolumn{#1}{c}{#2}}
\definecolor{Gray}{gray}{0.85}
\definecolor{LightCyan}{rgb}{0.88,1,1}
\definecolor{Cyan}{rgb}{1,0.9,0.3}
\definecolor{Magenta}{rgb}{1,0.5,1}

\newcolumntype{a}{>{\columncolor{Gray}}c}
\newcolumntype{b}{>{\columncolor{white}}c}

\begin{document}

\title{Perceptually Optimized Generative Adversarial Network for Single Image Dehazing}

\author{Yixin Du and Xin Li~\IEEEmembership{Fellow,~IEEE}

\thanks{Authors are with the Lane Department of Computer Science and Electrical Engineering, West Virginia University, Morgantown WV 26506-6109, USA}
\thanks{Corresponding author: Xin Li (xin.li@ieee.org)}}
\maketitle

\begin{abstract}
Existing approaches towards single image dehazing including both model-based and learning-based heavily rely on the estimation of so-called transmission maps. Despite its conceptual simplicity, using transmission maps as an intermediate step often makes it more difficult to optimize the perceptual quality of reconstructed images. To overcome this weakness, we propose a direct deep learning approach toward image dehazing bypassing the step of transmission map estimation and facilitating end-to-end perceptual optimization. Our technical contributions are mainly three-fold. First, based on the analogy between dehazing and denoising, we propose to directly learn a nonlinear mapping from the space of degraded images to that of haze-free ones via recursive deep residual learning; Second, inspired by the success of generative adversarial networks (GAN), we propose to optimize the perceptual quality of dehazed images by introducing a discriminator and a loss function adaptive to hazy conditions; Third, we propose to remove notorious halo-like artifacts at large scene depth discontinuities by a novel application of guided filtering. Extensive experimental results have shown that the subjective qualities of dehazed images by the proposed perceptually optimized GAN (POGAN) are often more favorable than those by existing state-of-the-art approaches especially when hazy condition varies.
\end{abstract}


\IEEEpeerreviewmaketitle
\section{Introduction}

Single image dehazing refers to the restoration of an image from its degraded observation under hazy conditions. To combat adversary conditions such as haze, physical modeling of the image degradation process has been extensively studied in the literature (e.g. \cite{middleton1957vision}, \cite{tan2008visibility}, \cite{narasimhan2002vision}). It is known that the process of light passing through a scattering medium such as atmosphere is characterized by the attenuation along the path of transportation. To make mathematical modeling tractable, it is often assumed that the fraction of light deflected and the distance traveled observe a linear relation. Such simplified assumption has led to the popular image formation model connecting observed hazy image with scene radiance (unknown target) and transmission map \cite{fattal2008single}. Based on such formation model, the problem of single image dehazing boils down to estimating the transmission map; and for this reason, many previous works on single image dehazing have focused on a model-based (e.g., uncorrelation principle \cite{fattal2008single}, dark channel prior \cite{he2011single}) or learning-based (e.g., dehazenet \cite{cai2016dehazenet}, multi-scale CNN \cite{ren2016single}) approach toward transmission map estimation.

We challenge this conventional wisdom by highlighting a few weaknesses of transmission-map-first approach. First, since the image formation model is based on simplified assumptions, it only represents an \emph{approximation} of the true in-scattering term in a full radiative transport equation \cite{fattal2008single}. The validity of this approximation becomes questionable in more realistic acquisition scenarios such as heavy haze and night environment. Therefore, the effectiveness of using a transmission map to recover scene radiance might deteriorate as hazy condition varies. Second, the transmission-map-first approach suffers from the potential \emph{error propagation} - i.e., any error in the estimated transmission map could have catastrophic impact on the recovered scene radiance. Surprisingly, this issue of error propagation has not been addressed in the open literature to the best of our knowledge. In previous works (e.g., \cite{he2011single},\cite{ren2016single}), only a small positive constant is added to the denominator for improving numeral stabilities of solution algorithms. Third, with the estimation of transmission map involved as an intermediate step, it becomes difficult to conduct end-to-end optimization especially from the \emph{perceptual} point of view.

\begin{figure}
        \includegraphics[width=\linewidth]{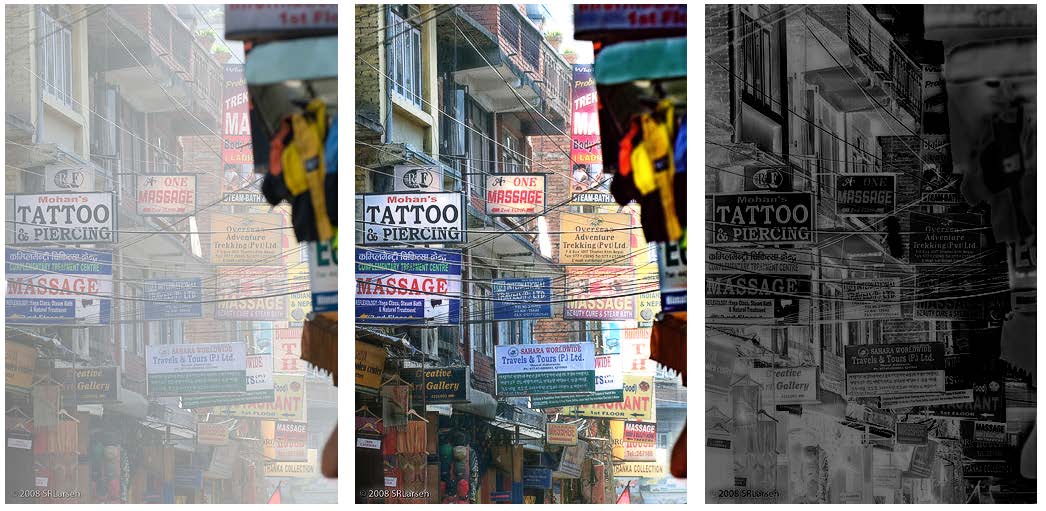}
        \caption{Haze removal without the necessity of estimating a transmission map. From left to right: input(hazy), output (dehazed) and residual (the difference between input and output).}
        \label{fig_intro}
\end{figure}

\begin{figure*}
\centering
\includegraphics[width=\linewidth]{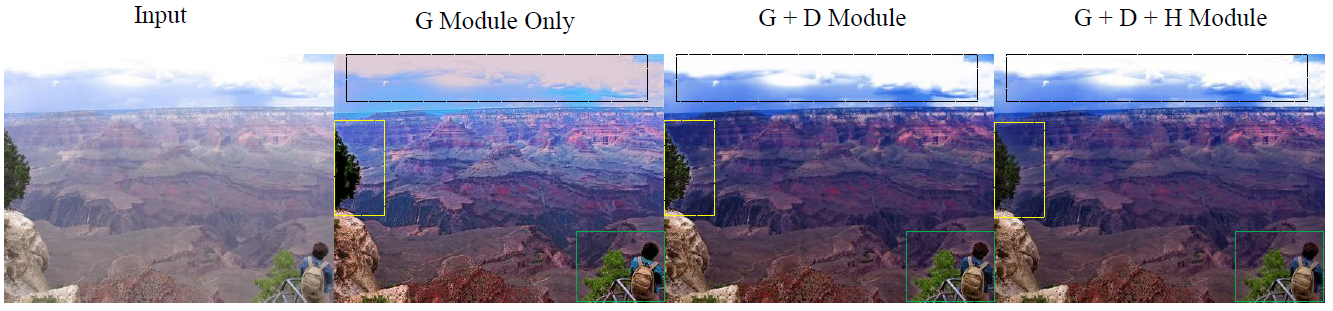}
\caption{Perceptual optimization achieved by the proposed GAN including a generative network \emph{G}, a discriminative network \emph{D}, and a post-processing module \emph{H}.}
\label{fig_intro_comp}
\end{figure*}

In this paper we advocate a \emph{direct} deep learning-based approach toward single image dehazing without estimating transmission map at all (as shown in Fig. \ref{fig_intro}) and capable of end-to-end perceptual optimization. Our direct approach is motivated by a flurry of most recent advances in the field of deep learning including deep residue networks \cite{he2016deep}, \cite{zhang2017beyond},\cite{tai2017image} and generative adversarial networks (GAN) \cite{goodfellow2014generative},\cite{denton2015deep},\cite{ledig2016photo}. The main contributions of this work are summarized by the three components as shown in Figure~\ref{fig_intro_comp}.

$\bullet$ \emph{Generative network}. Inspired by the analogy between denoising and dehazing, we propose to directly learn a nonlinear mapping from the space of degraded images to that of haze-free ones via deep residual network \cite{he2016deep}. Since our approach does not rely on estimating transmission maps as an intermediate step, it can work with a variety of hazy conditions (both heavy and light) no matter whether the image formation model holds or not. Moreover, by feeding the output of the network as the input, we can obtain a \emph{recursive} extension of deep residual learning; in other words, haze-free images can be viewed as the fixed-point \cite{goebel1972fixed} of our generative network.

$\bullet$ \emph{Discriminative network}. As mentioned above, it is often difficult to address the issue of perceptual quality in transmission-map-first approaches. In this work, we propose to leverage the success of generative adversarial networks (GAN) from image synthesis \cite{denton2015deep} and super-resolution \cite{ledig2016photo} to single image dehazing. Based on our discriminative network, we propose to optimize the perceptual quality of dehazed images by introducing an \emph{adaptive} loss function. Adaptive weights in our loss function are conceived to facilitate perceptual optimization of GAN-based dehazing when hazy condition varies.

$\bullet$ \emph{Post-processing module}. In view of the tendency of producing various artifacts in dehazed images (e.g., color shifting \cite{chen2016robust} and halo-like \cite{gibson2012investigation}), we propose to remove the unpleasant artifacts by a novel application of guided filtering \cite{he2010guided}. More specifically, the hazy image will serve as a guidance for correcting the recursively-learned residue image. The effectiveness of such guided filtering based post-processing on suppressing various artifacts has been verified especially around large scene depth discontinuities.

When compared with previous approaches, our Perceptually Optimized GAN (POGAN) can be trained in an end-to-end fashion because it bypasses the unnecessary step of transmission map estimation. By explicitly addressing the issue of perceptual quality, our POGAN can be optimized for both indoor and outdoor scenes and under a variety of haze conditions (heavy vs. light). We have conducted extensive experimental studies with respect to both synthetic and real-world images. In our study, we have compared both subjective and objective visual quality of dehazed images and found that our POGAN often performs favorably against other state-of-the-art approaches especially in terms of subjective evaluation. Restored images by this work are often the most faithful reproduction of original images with respect to color vividness and fine structural details.

\section{Related Works}

In model-based approaches, estimating transmission map is often necessary which requires some haze-relevant prior. In \cite{he2011single}, a Dark Channel Prior (DCP) was proposed based on the observation that any local patch would contain some pixels of low intensity values in at least one color channel of a haze-free image. In some scenario (e.g., when an image contains a lot of sky pixels), the performance of DCP-based dehazing degrades rapidly. In \cite{nishino2012bayesian}, a factorial Markov Random Field model was adopted to jointly estimate the scene albedo and depth. The authors reported accurate factorization on challenging scenes of the proposed method. The main disadvantage is that it tends to produce over-saturated images. In \cite{meng2013efficient}, the inherent boundary constraint of the transmission function was exploited during the estimation of transmission maps, which has been called contextual regularization. In nonlocal image dehazing \cite{berman2016non}, distance map and haze-free images are jointly estimated from haze-lines characterizing the linear-spreading structure of pixels within a given cluster in the color space. One disadvantage of this nonlocal approach is that it assumes a fixed distribution of 3D lines from the air light which limits its capability of describing an actual scene~\cite{berman2017air}.

Inspired by the success of deep learning in image super-resolution~\cite{dong2014learning,dong2016image} and denoising~\cite{xie2012image,zhang2017beyond}, learning-based approaches have also been proposed for singe image dehazing in recent years ~\cite{ren2016single,cai2016dehazenet,li2017aod}. In \cite{ren2016single},  a multi-scale convolutional neural network (MSCNN) was proposed to estimate transmission map from an input hazy image. It consists of a coarse-scale net which aims at globally predicting the holistic transmission map of a scene, and a fine-scaled network to further refine the transmission map estimation locally. In \cite{cai2016dehazenet}, DehazeNet was proposed with network layers specially tailored to fit assumptions/priors in the scenario of image dehazing. Similar to MSCNN, DehazeNet is also an indirect learning based dehazing approach which recovers transmission map first. Most recently, the so-called All-in-One Dehazing (AOD)\cite{li2017aod} further develops this line of idea by unifying the estimation of transmission map and global atmosphere light into one module called K-estimation leading to the current state-of-the-art performance.

\section{Proposed Approach}

To facilitate the discussion, we start from a brief review of the simplified degradation model for a hazy image (note that our approach does not rely on this model):

\begin{equation}
\matr I(x) = \matr J(x)t(x)+\matr A(1-t(x)),
\label{equ:haze_formation}
\end{equation}
where $\matr A$ is the global atmospheric light, $t(x)$ is the medium transmission map characterizing the amount of attenuation, $\matr J$ is the scene radiance (target of restoration), and $\matr I$ is the observed hazy image. On the right-hand side of Eq. \eqref{equ:haze_formation}, $\matr J(x)t(x)$ represents the term of \textit{direct attenuation} and $\matr A(1-t(x))$ is called \textit{airlight} \cite{middleton1957vision,tan2008visibility,he2011single}. Direct attenuation accounts for decay of scene radiance through the transmission medium; and airlight is attributed to atmospheric scattering~\cite{he2011single}. The transmission map $t(x)$ is related to the scene depth by:
\begin{equation}
t(x) = e^{-\beta d(x)},
\label{equ:transmission}
\end{equation}
where $\beta$ is the scattering coefficient of the atmosphere and $d(x)$ is the depth of the scene at location $x$. It is easy to see that the amount of attenuation depends on the distance from the scene point to the camera~\cite{he2011single} - the larger the distance, the stronger the attenuation. Additionally, the global atmospheric light would interfere the process of image degradation, which make image dehazing more challenging in night-time than in day-time. In previous approaches, estimating transmission map $t(x)$ or its variation (e.g., K-estimation in \cite{li2017aod}) is a necessary step for both model-based and learning-based dehazing.

\subsection{Generative Network via Recursive Deep Residue Learning}

In this work, we propose to take a direct approach of learning a nonlinear mapping from $\matr I(x)$ to $\matr J(x)$ based on an analogy between dehazing and denoising. If we reformulate Eq. \eqref{equ:haze_formation} as follows:
\begin{equation}
\begin{split}
\matr I(x) &= \matr J(x) + (\matr A-\matr J(x))(1-t(x))\\
&= \matr J(x) + \matr r(x)
\end{split}
\label{equ:reformulation}
\end{equation}
where $\matr r(x)=(\matr A-\matr J(x))(1-t(x))$ can be interpreted as a \emph{structured} error term characterizing the nonlinear signal-dependent degradation associated with the hazy effect. Such reformulation enables us to connect dehazing with the widely-studied problem of denoising - i.e.,
\begin{equation}
\matr I(x) = \matr J(x) + \matr w(x)
\label{equ:awgn}
\end{equation}
where $\matr I(x), \matr J(x)$ denote noisy and clean images respectively, the additive noise term $\matr w(x) \sim N(0,\sigma_w^2)$ is often assumed to be white Gaussian in the denoising literature. By comparing Eq. \eqref{equ:awgn} and Eq. \eqref{equ:reformulation}, we observe the apparent analogy between two error terms - conceptually speaking, if a residue network \cite{he2016deep} can learn white Gaussian noise from degraded images, it can also learn a structured one.

The analogy between dehazing and denoising also leads to a nonlinear optimization of residue learning-based dehazing via iterative regularization \cite{osher2005iterative}. In image denoising, suppose a regularized estimate of clean image from noisy observation $\matr I(x)$ is given by a nonlinear mapping $\Phi^{-1}(\matr I(x))$ and the error is denoted by $\matr e(x)=\matr I(x)-\Phi^{-1}(\matr I(x))$. If $\matr e(x)$ is already white Gaussian, then we are done; otherwise (i.e., $\matr e(x)$ still contains some leftover image structures), a simple strategy of further improvement is to feed the denoised image $\Phi^{-1}(\matr I(x))$ back to the denoising algorithm and see if the new error is closer to zero (when a clean image is the fixed point, the residue goes to zero). Similarly, we can recursively feed a dehazed image back to the input of the proposed residue network - if the haze-free image is the fixed-point, the learned residue should asymptotically goes to zero.

\begin{figure}[!t]
\centering
\includegraphics[width=3.5in]{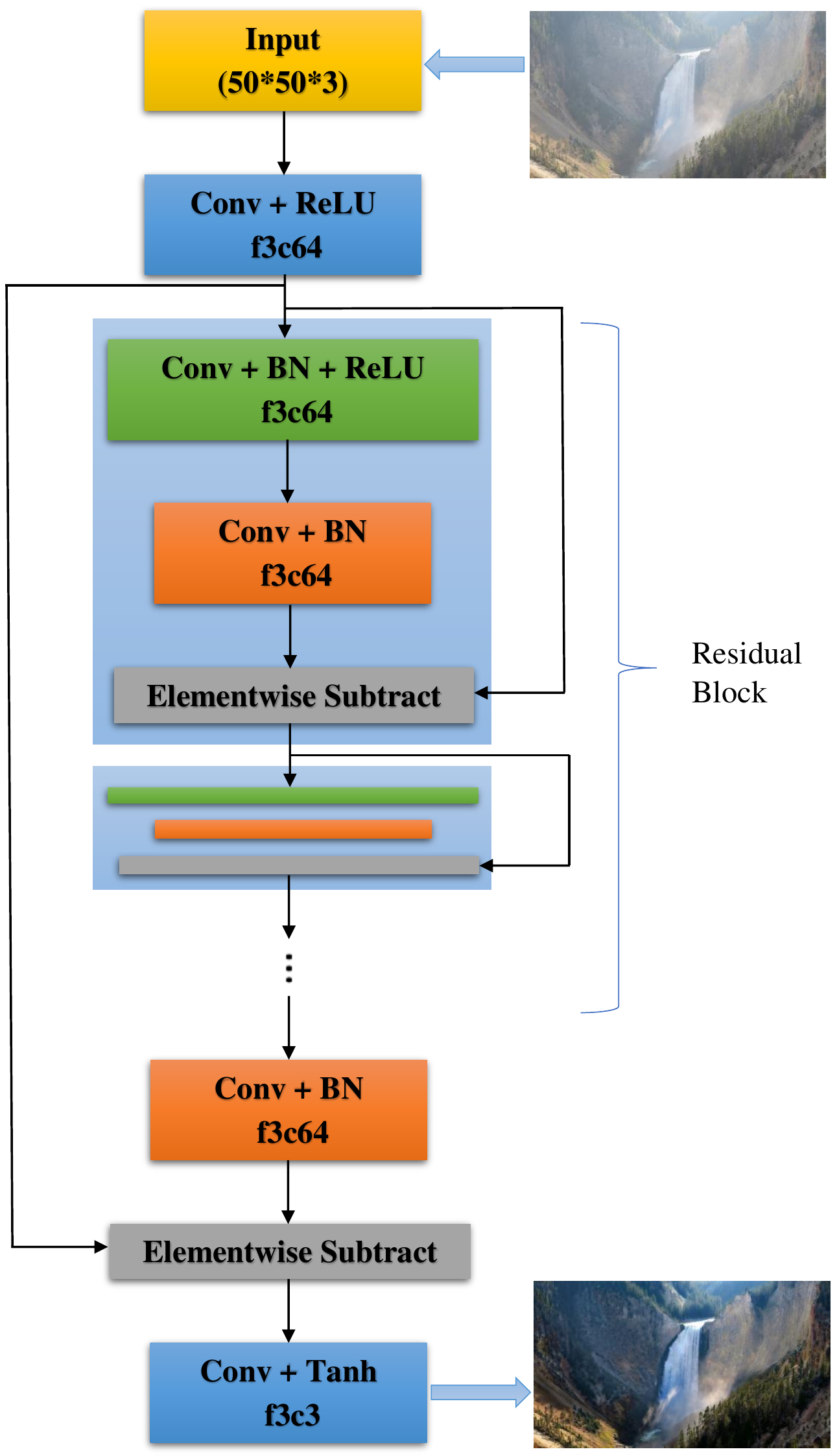}
\caption{Architecture of a deep residual network with corresponding filter size (f) and the number of feature channels (c).}
\label{fig_drl}
\end{figure}

Figure~\ref{fig_drl} shows the architecture of our recursive deep residual learning module with corresponding filter size (f) and the number of feature channels (c). The module takes a hazy patch ($50\times50\times3$) as input, followed by Convolution (Conv) and Rectified Linear Unit (ReLU) layer with 64 feature channels and $3\times3$ filter size. The residual block includes 16 sub-blocks. Each sub-block is composed of Conv, Batch Normalization (BN), Relu, Conv, and a Elementwise (Elti) Subtraction layer. The Elti layer takes the input from last sub-block, subtracts the residual recovered in the current sub-block, and return a less-hazy patch in a progressive manner. The last Elti layer performs a pixel-wise subtraction of the input and output of residual block followed by another Conv layer with a hyperbolic tangent activation function.

\subsection{Discriminative Network with Perceptually Optimized Loss Function}

As mentioned above, it is usually difficult to address the issue of visual quality assurance in previous transmission-map-first approaches. Bypassing the estimation of transmission map makes it possible to leverage the idea of generative adversarial networks (GAN) from image synthesis \cite{denton2015deep} and super-resolution \cite{ledig2016photo} to image dehazing.
The basic idea behind GAN is to introduce a discriminative network as a judge telling whether the output of generative network is real or fake. Under the context of dehazing, $G$ produces dehazed image patches and  $D$ classifies them as dehazed (fake) and haze-free (real). The goal of adversarial learning is for $G$ to produce dehazed patches that can fool $D$ (dehazed patches are visually indistinguishable from haze-free ones). We have followed the discriminative architecture guidelines proposed in \cite{ledig2016photo,radford2015unsupervised} which contains Conv layers with 64, 128, 256, and 512 channels. The last Conv layer is followed by two dense layers and a sigmoid activation function to classify the dehazed and haze-free patches.

We have adopted the perceptual loss function proposed in~\cite{ledig2016photo} which is a weighted sum of MSE, VGG loss, and adversarial loss respectively:
\begin{equation}
l = w_1l_{MSE} + w_2l_{VGG} + w_3l_{Adv},
\label{equ:loss}
\end{equation}
where $w_i$ (\emph{i=1,2,3}) controls the weight of each term. The pixel-wise MSE loss is given by:
\begin{equation}
l_{MSE} = \frac{1}{WH}\sum_{x=1}^{W}\sum_{y=1}^{H}(\matr J_{x,y}^{*}-\matr J_{x,y}^{G})^2,
\label{equ:loss_mse}
\end{equation}
where $\matr J^{*}$ denotes the ground truth (haze-free image) and $\matr J^{G}$ denotes the dehazed one by the $G$ module. The VGG loss computes the Euclidean distance between the feature maps of $\matr J^{*}$ and $\matr J^{G}$:
\begin{equation}
l_{VGG} = \frac{1}{W_{i,j}H_{i,j}}\sum_{x=1}^{W_{i,j}}\sum_{y=1}^{H_{i,j}}(\phi_{i,j}(\matr J^{*})_{x,y}-\phi_{i,j}(\matr J^{G})_{x,y})^2,
\label{equ:loss_mse}
\end{equation}
where $W_{i,j}$ and $H_{i,j}$ denotes the dimensions of extracted feature maps. Finally, the adversarial loss can be written as:
\begin{equation}
l_{Adv} = \sum_{n=1}^N -\log{(\matr J^{D})},
\label{equ:loss_adv}
\end{equation}
where $\matr J^{D}$ is the probability that the reconstructed image is haze-free.

To the best of our knowledge, previous GANs mostly use loss functions with fixed weights for each module such as~\cite{goodfellow2014generative,denton2015deep,radford2015unsupervised, huang2017stacked,liu2016coupled}. Toward the objective of perceptual optimization, we propose an adaptive perceptual loss function tailored to fit the severity of haze in an image. The rationale is that the process of dehazing has to deal with various uncertainty factors such as direct attenuation and airlight in the image degradation model. Since the attenuation term $\matr J(x)t(x)$ dominates the thickness of haze during the degradation, it is natural to adaptively choose the weights of loss function based on the attenuation term. That is, we can adjust $w_1$, $w_2$, and $w_3$ based on the amount of attenuation controlled by $\beta$ (large $\beta$ corresponding to heavy haze). More specifically, we propose to use larger $w_1$ under heavy haze situation (i.e., more emphasis on haze removal) and larger $w_3$ under light haze condition (i.e., more emphasis on quality assurance).

\subsection{Post-processing Module for Halo Removal}

It has been widely recognized that dehazed images have the tendency of producing various halo-like artifacts (e.g., ringing reduction \cite{gibson2012investigation}, anti-halation enhancement \cite{li2017aod}, block halo suppression~\cite{zhai2015single}). We have also empirically observed that the proposed GAN-based dehazing sometimes suffer from noticeable halo-like artifacts especially around the areas of large depth discontinuities (i.e., rapid change of attenuation) as shown in Figure~\ref{fig_hb}. We have also found that using larger filters (e.g., $5 \times 5$, $7 \times 7$, or a combination of filters with different sizes) tend to make the artifacts more serious (the so-called block halo problem \cite{zhai2015single}).
\begin{figure*}
\includegraphics[width=1\linewidth]{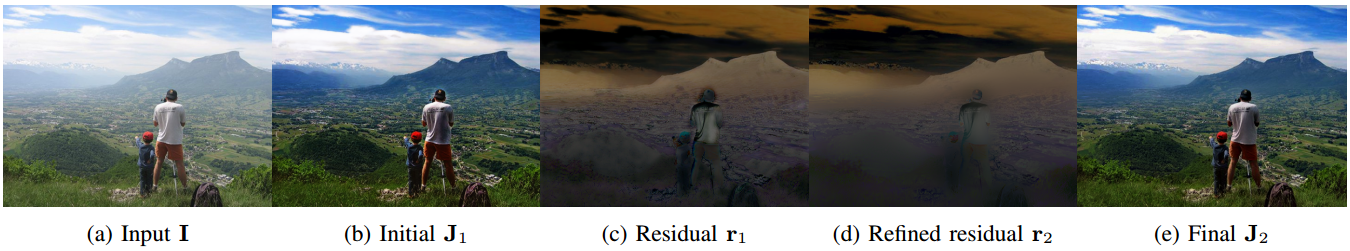}
                \label{fig_he}
        \caption{The proposed post-processing module for halo suppression. The initial dehazed image $\matr J_1$ is the input. We first obtain $\matr r_1$ via elemental-wise subtraction $Elti(\matr I,~\matr J_1)$. Then refined residual $\matr r_2$ is obtained by applying guided filtering to $\matr r_1$ ($\matr I$ is the guidance). Finally, $\matr J_2$ is recovered from $Elti(\matr I,~\matr r_2)$.}
        \label{fig_halo}
\end{figure*}

To suppress the potential halo-like artifacts, we propose to include a guided-filtering based postprocessing module. Guided filtering was first proposed in~\cite{he2010guided} and its effectiveness on refining the estimated transmission map has been well documented in the literature (e.g., \cite{pang2011improved}). Here we suggest a novel application of this powerful tool into refining the residual map as a post-processing strategy. Guided filtering assumes the following linear relationship between the guidance $I$ and output $q$:
\begin{equation}
q_i=\alpha_k I_i+b_k, \forall i \in \omega_k,
\label{equ:qAssm}
\end{equation}
where $(a_k,b_k)$ are some constant linear coefficients. To compute these coefficient, one needs to minimize a cost function characterizing the difference between $q$ and the input $p$ in a window $\omega_k$~\cite{he2010guided}:
\begin{equation}
E(a_k,b_k)=\sum_{i \in \omega_k} ((a_k I_i + b_k - p_i)^2 + \epsilon \alpha_k^2)
\label{equ:pAssm}
\end{equation}

\begin{figure}
        \begin{subfigure}{\linewidth}
                \includegraphics[width=1\linewidth]{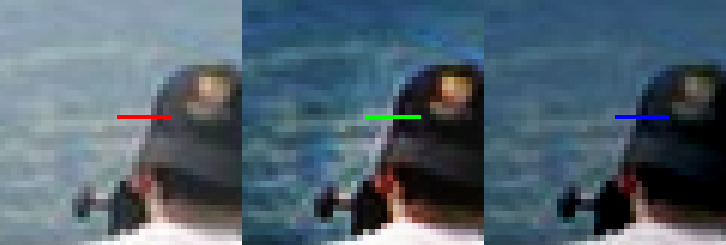}
                \caption{From left to right: $\matr I$, $\matr J_1$, and $\matr J_2$}
                \label{fig_ca}
        \end{subfigure}
        \begin{subfigure}{\linewidth}
        		    \centering
                \includegraphics[width=1.8in]{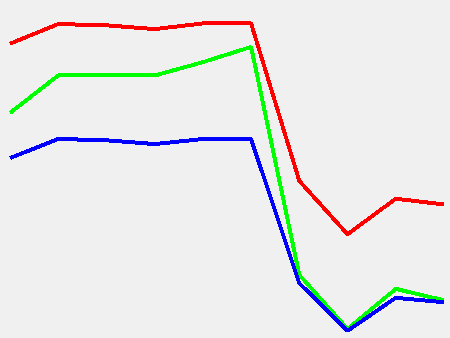}
                \caption{1-D illustration for halo removing.}
                \label{fig_cb}
        \end{subfigure}
        \caption{Visual comparison of zoomed portions in Fig. \ref{fig_halo} and the corresponding 1D intensity profiles.}
        \label{fig_haloComp}
\end{figure}

As shown in Figure~\ref{fig_halo}, our novel application of guided filtering into halo removal consists of three steps. First, We obtain raw residual $\matr r_1$ via elemental-wise subtraction $Elti(\matr I,~\matr J_1)$ where $\matr J_1$ is recovered by applying $G$ and $D$ module on the hazy image $\matr I$. Second, the \emph{refined} residual $\matr r_2$ is obtained by applying a guided filter using $\matr I$ as the guidance and $\matr r_1$ as the input image. Finally, refined image estimation $\matr J_2$ is recovered from $Elti(\matr I,~\matr r_2)$. To better illustrate how the proposed postprocessing module works, we have used a toy example as shown in Figure~\ref{fig_haloComp}. Figure~\ref{fig_ca} shows the zoomed comparison of $\matr I$, $\matr J_1$, and $\matr J_2$ around the head region (where large scene depth discontinuity occurs) respectively; and figure~\ref{fig_cb} shows the corresponding 1-D plot of intensity pixels. It can be clearly observed how the overshooting estimate by our generative network gets corrected by guided filtering.

\section{Experimental Results}

\subsection{Datasets and Implementation Details}

\begin{figure}
        \includegraphics[width=\linewidth]{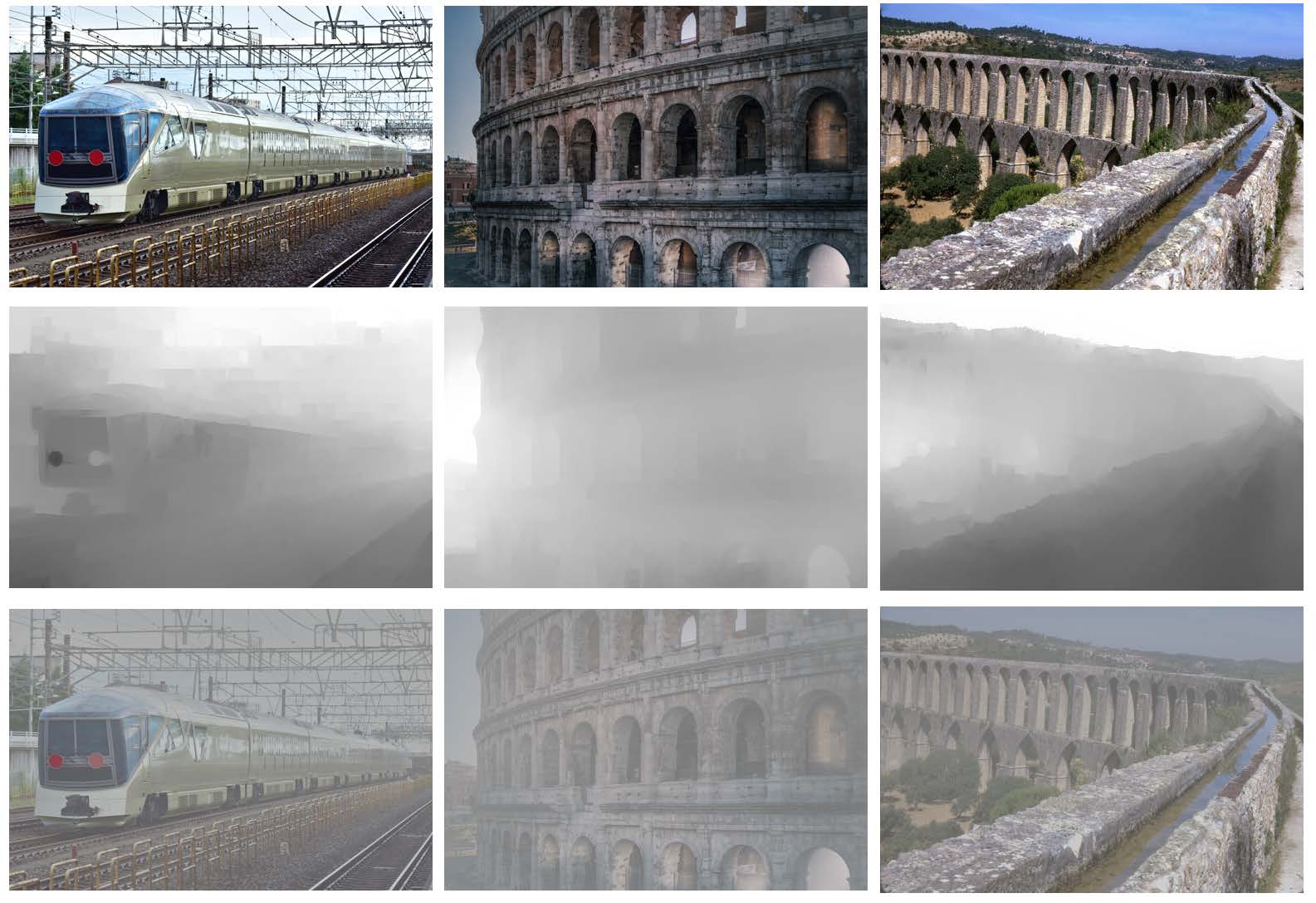}
        \caption{Creation of training dataset. From top to bottom row: original image from DIV2k~\cite{agustsson2017ntire} dataset, depth map computed using~\cite{liu2016learning}, and the hazy images generated by Eq. \eqref{equ:haze_formation}. }
        \label{fig:div2k}
\end{figure}

Preparation of training data plays an important role in deep learning-based approaches. Previous works such as MSCNN \cite{ren2016single} and AOD \cite{li2017aod} have used the NYU-Depth V2 \cite{silberman2012indoor} dataset where color and depth images are captured by Microsoft Kinect. In view of the limited image quality of the NYU-Depth V2 dataset, we have taken 799 image from the DIV2K \cite{agustsson2017ntire} dataset with high quality images (originally constructed for image super-resolution). To obtain the corresponding depth images, we have borrowed a deep CNN-based approach of learning depth images from single monocular images \cite{liu2016learning}. Figure~\ref{fig:div2k} shows some examples of the learned depth maps for the preparation of training dataset. To simulate synthetic hazy images, the following parameters are used in our experiments: we have randomly selected attenuation parameter $\beta \in \lbrace0.5,0.6,0.7,0.8,0.9,1.0,1.1,1.2,1.3,1.4,1.5\rbrace$ since any value beyond this range could lead to unrealistic haze (too thin or too heavy) and unwanted noise amplification~\cite{ren2016single}. For each of the RGB channel, atmospheric light $A$ is chosen uniformly within the range of $\left[ 0.7, 1.0\right]$. 

The test dataset consists of both synthetic and real-world hazy images. Similar to the previous work \cite{li2017aod}, the synthetic test data contains 100 images from the DIV2K dataset as \textbf{ImageSet A} and 21 images from the Middlebury Stereo Datasets \cite{scharstein2003high} as \textbf{ImageSet B}; additionally, we pick another 31 real-world images as \textbf{ImageSet C}. During the training process, the weights of each convolution layers are randomly initialized by Gaussian variables. The patch size is $50 \times 50$; the number of epochs is set to 100; the learning rates for the first 50 and the remaining 50 epochs are set to 0.001 and 0.0001 respectively. We have selected Adam optimizer with Beta1 parameter being 0.9. The network is implemented using TensorFlow and trained on a PC with an Intel i7-4790k processor and a Nvidia GeForce Titan GPU.

\subsection{Effectiveness of Each Module}

\begin{figure}
        \includegraphics[width=\linewidth]{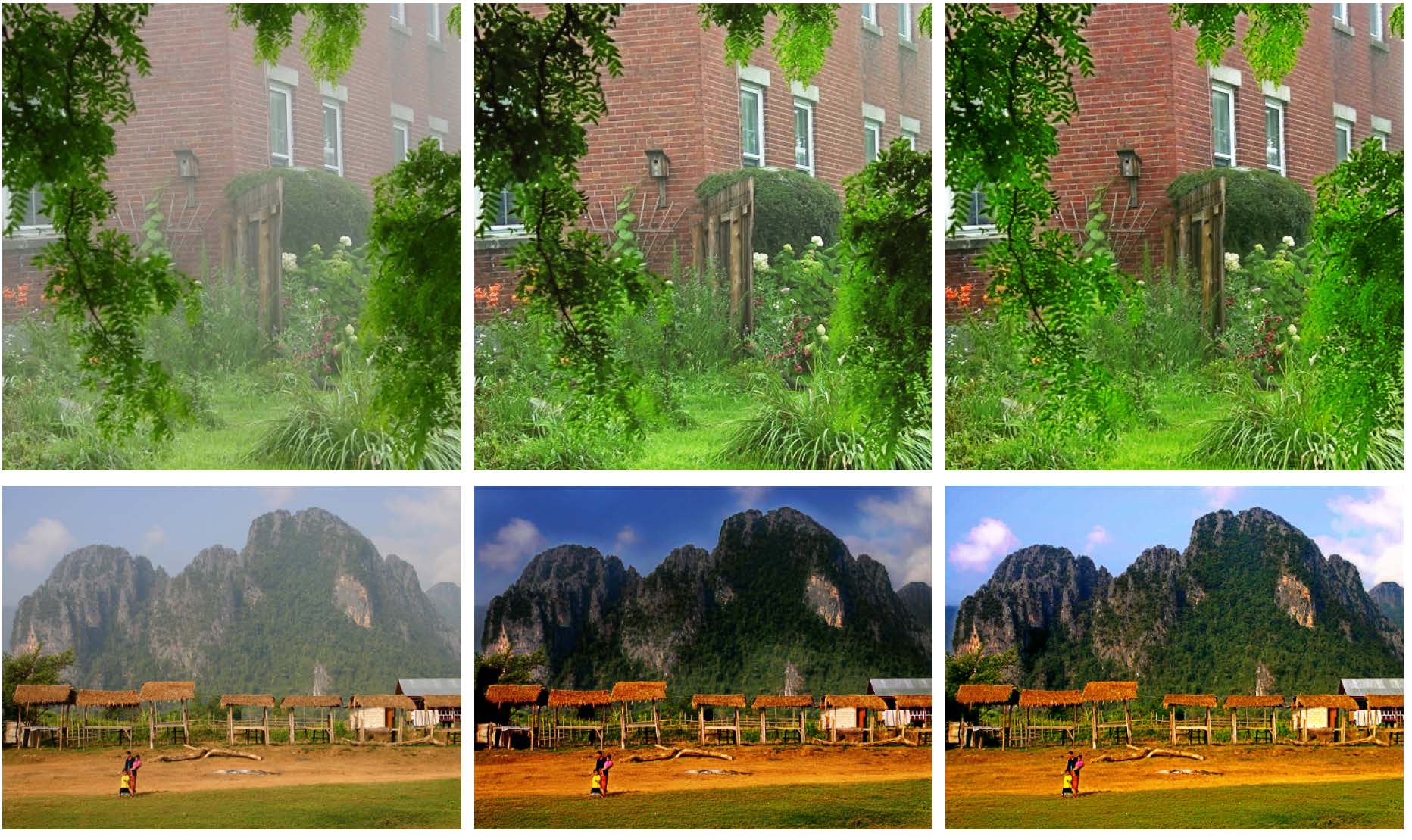}
        \caption{Discriminative network improves the visual quality of dehazed images. Left: input; middle: dehazing with G module optimized using MSE only; right: dehazing with both G and D module optimized for human perception.}
        \label{fig:gVSgd}
\end{figure}

We have designed a series of comparative studies to facilitate the illustration of each module in the proposed approach. First, we want to demonstrate the effectiveness of introducing a \emph{discriminative} network on visual quality assurance. Figure~\ref{fig:gVSgd} shows the dehazing results on a pair of real-world images without and with a discriminative network. It can be observed that dehazing with a generative network only tends to remove haze over aggressively, especially in the background where there is heavy haze. The undesirable consequence is that some part of the foreground (e.g., trees and mountains) becomes unnaturally dark. By contrast, the inclusion of a discriminative network makes the dehazed images visually more pleasant as shown in the right column of Figure~\ref{fig:gVSgd}. This is due to the perceptual loss function helps to ensure that the dehazed images are as close as possible to real haze-free images.

\begin{figure}
        \includegraphics[width=\linewidth]{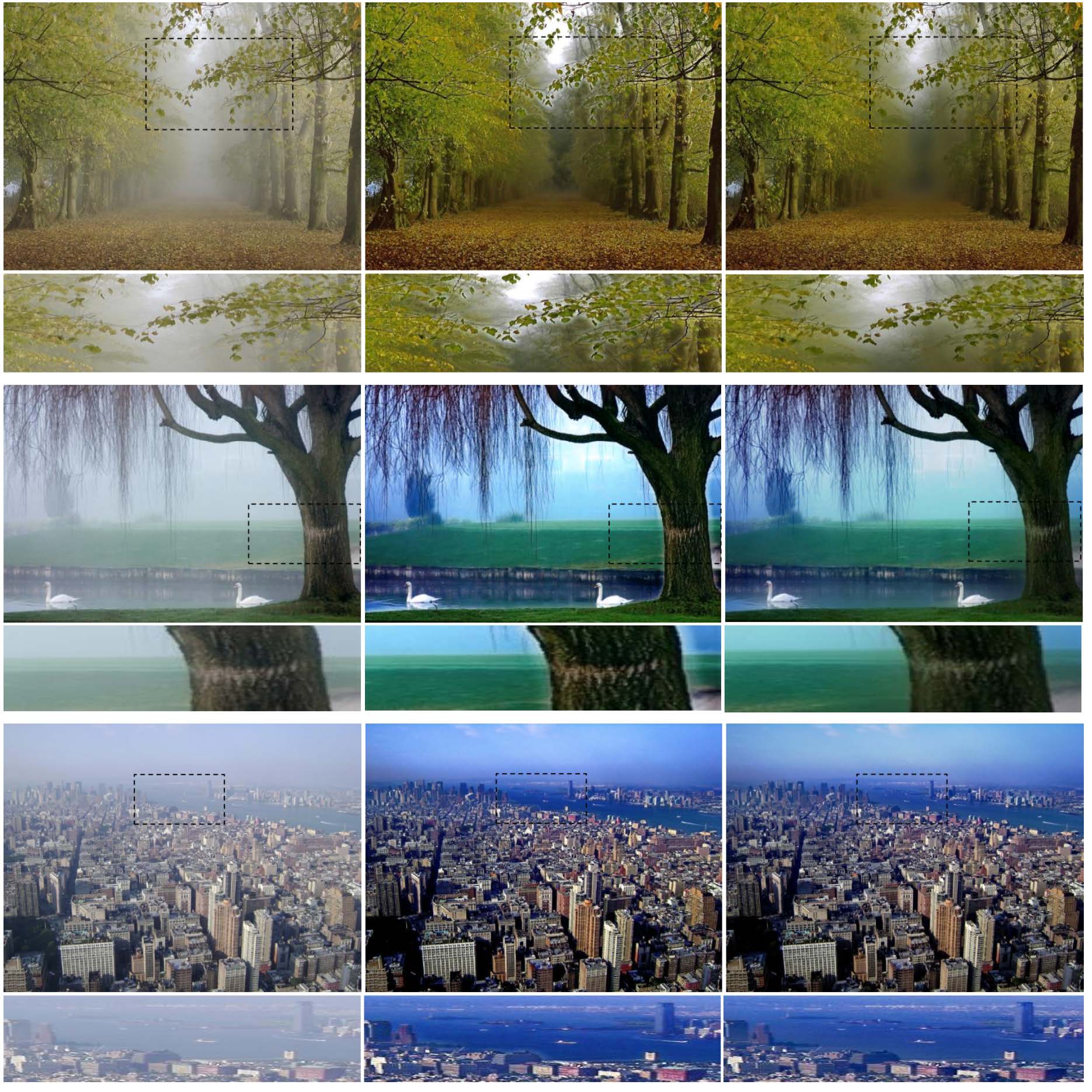}
        \caption{Post-processing module improves the visual quality of dehazed images. From left to right: input, dehazing results without and with the post-processing module.}
        \label{fig:gdVSgdh}
\end{figure}

Second, we need to justify the effectiveness of the proposed \emph{post-processing} module. Even though with a discriminative network, we have observed that noticeable halo artifacts could occur around areas with large scene depth discontinuities (e.g., within 4-8 pixels away from the boundary between foreground and background). Meantime, the larger the filter size, the more serious the halo artifacts become which agrees with the observation made in~\cite{zhai2015single}.  Figure~\ref{fig:gdVSgdh} shows that the comparison of dehazed images before and after the proposed post-processing module. It can be clearly seen that undesirable halo artifacts in highlighted dashed areas have been successfully suppressed after post-processing.

\begin{figure}[!t]
\centering
\includegraphics[width=3.5in]{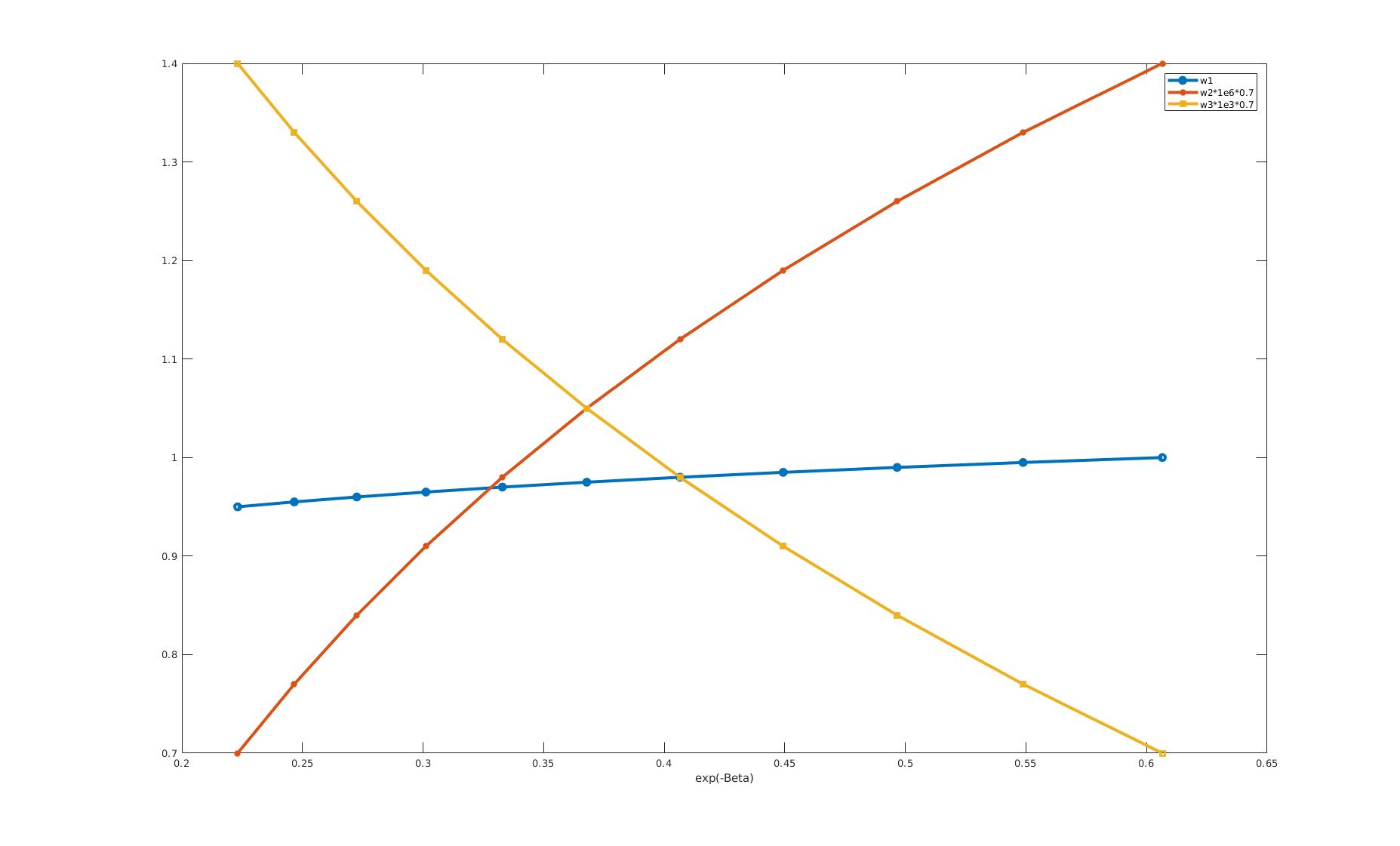}
\caption{The adjustment of weights in Eq. \eqref{equ:loss} based on the severity of attenuation ($w_1,w_2,w_3$ correspond to blue, red and yellow respectively).}
\label{fig_w}
\end{figure}

\begin{figure}
        \includegraphics[width=\linewidth]{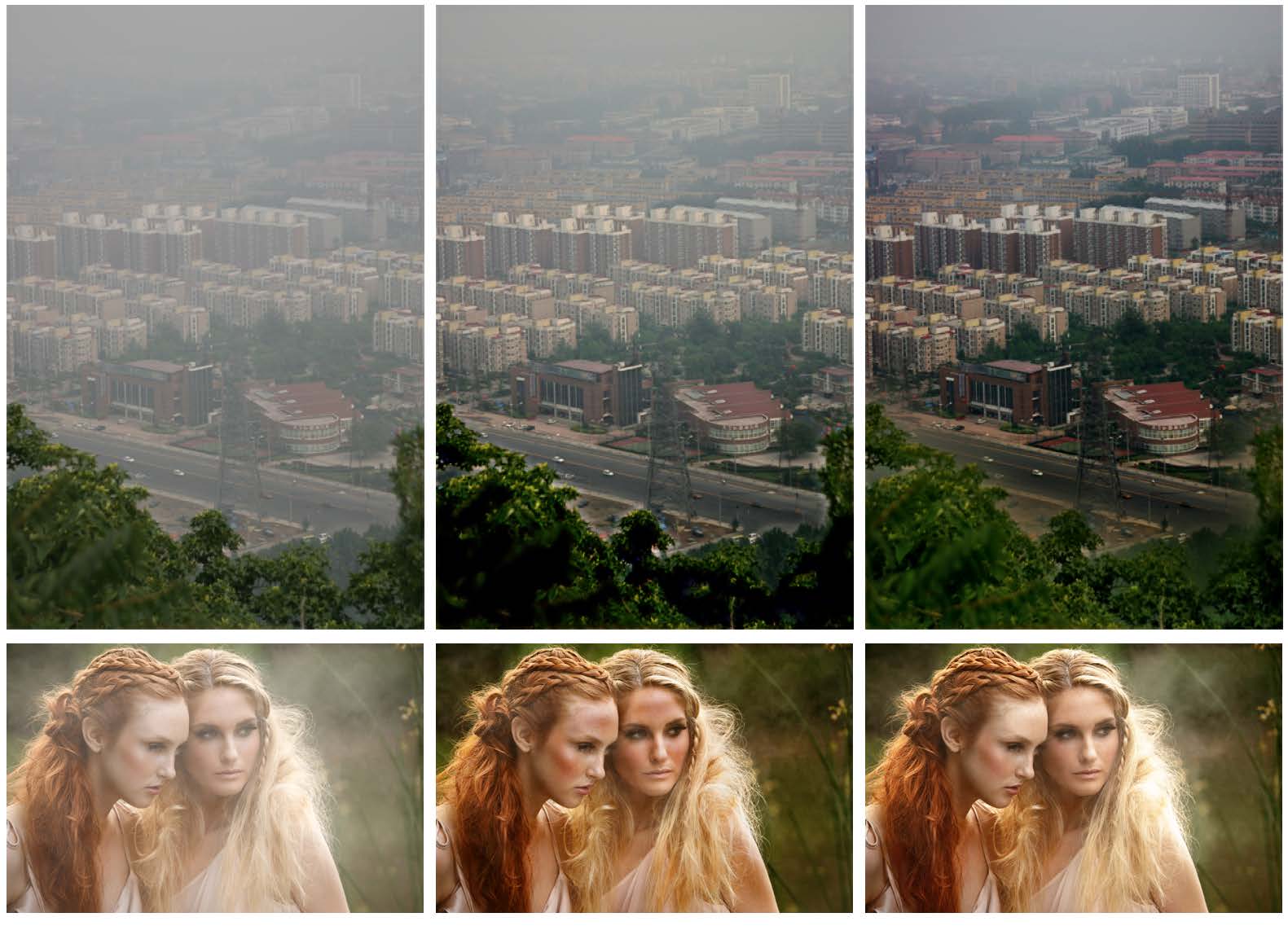}
        \caption{Benefit of adaptive perceptual loss function in GAN-based dehazing. From left to right: input, dehazing results of GAN with fixed weights, and dehazing results of GAN with adaptive weights (top: heavy haze; bottom: light haze).}
        \label{fig:fixVSaa}
\end{figure}

Third, we aim at illustrating the benefit of \emph{adaptive} perceptual loss function in GAN-based dehazing. In our experiments, we increase $w_1$  from 0.95 to 1 and $w_2$ from 0.000001 to 0.000002 respectively; and decrease $w_3$ from 0.002 to 0.001 as the attenuation parameter $\beta$ varies within its operational range (as shown in Figure~\ref{fig_w}). Such adaptive weight setting is compared against a fixed setting ($w_1=1, w_2=0.000001, w_3=0.002$). Figure~\ref{fig:fixVSaa} shows the comparison of dehazed images between fixed and adaptive weights under varying haze conditions. It can be verified that the proposed strategy of adaptive weights are capable of more effectively removing heavy haze while preserving the visual quality under light haze conditions (i.e., to achieve the objective of perceptual optimization).

\subsection{Comparison against State-of-the-Art on Synthetic Images}

We have compared our POGAN-based image dehazing with several state-of-the-art dehazing methods including: Dark Channel Prior (\textbf{DCP})~\cite{he2011single}, Non-Local Image Dehazing (\textbf{NLD})~\cite{berman2016non}, Boundary Constrained Context Regularization (\textbf{BCCR})~\cite{meng2013efficient}, Multi-Scale CNN (\textbf{MSCNN})~\cite{ren2016single}, \textbf{DehazeNet}~\cite{cai2016dehazenet}, and \textbf{AOD}~\cite{li2017aod}. The first three methods, including DCP, BCCR and NLD, are state-of-the-art model-based methods, and the last three are leading learning-based methods. Two objective image quality metrics have been used in our comparison: Peak Signal-to-Noise Ratio (PSNR) and Structural Similarity (SSIM). We have conducted experiments on both ImageSet A and B with $\beta \in \lbrace0.75, 1, 1.25, 1.5\rbrace$ and $A \in \lbrace 0.7, 0.8, 0.9, 1\rbrace$. Table \ref{tab:sync_AB} shows the average PSNR and SSIM comparison results on ImageSet A and B respectively. Overall, it can be observed that DCP and AOD respectively outperform others on ImageSet A and B respectively; however, only ours can perform equally well on both datasets. In fact, our PSNR/SSIM performance only slightly falls behind the best method on both data sets.

\begin{figure}[!h]
        \includegraphics[width=\linewidth]{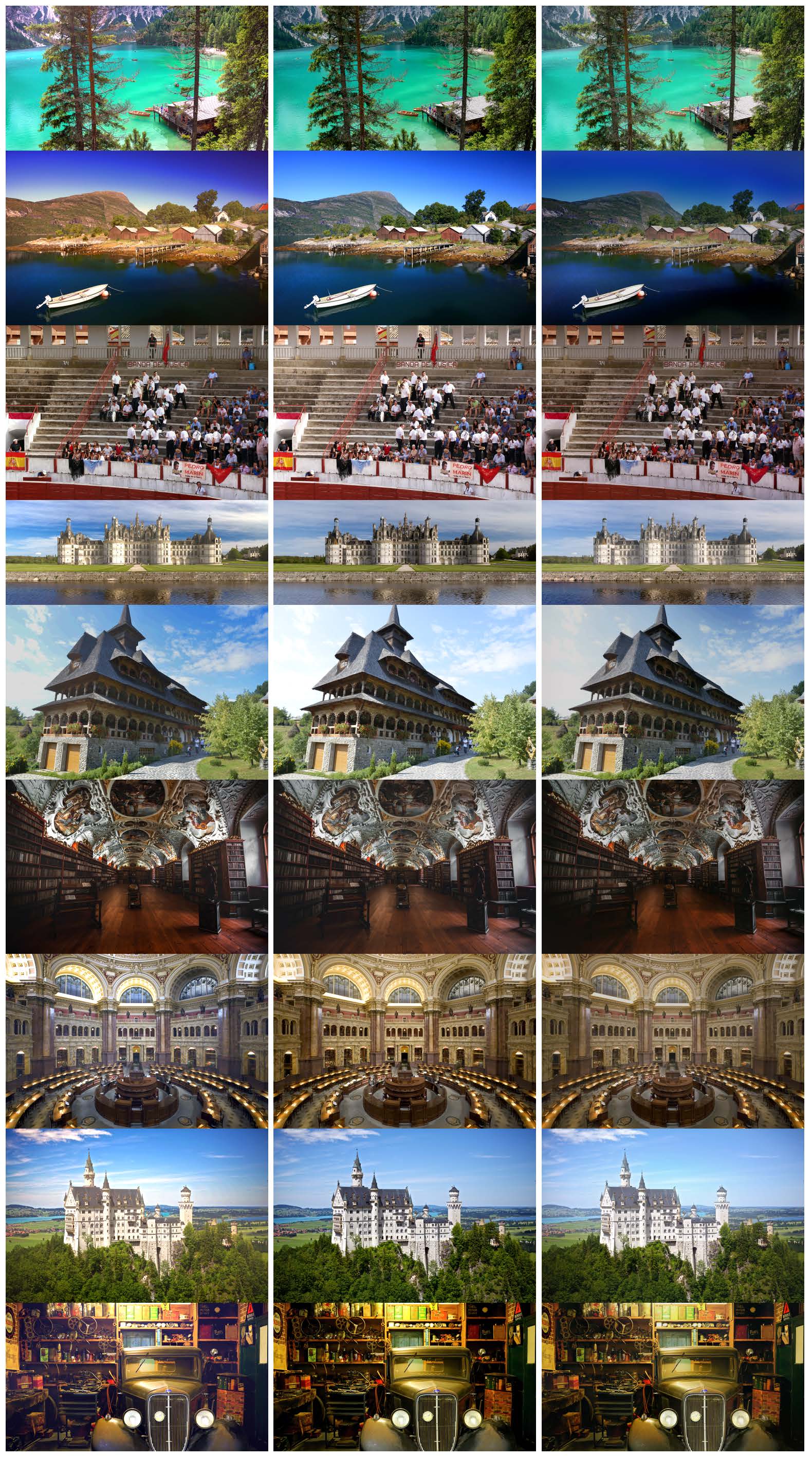}
        \caption{Dehazing results on ImageSet A (full-size comparison can be found in the supplementary material). Left: DCP results~\cite{he2011single}. Middle: original images. Right: outputs of POGAN.}
        \label{fig:imagesetA}
\end{figure}

\begin{figure}
        \includegraphics[width=\linewidth]{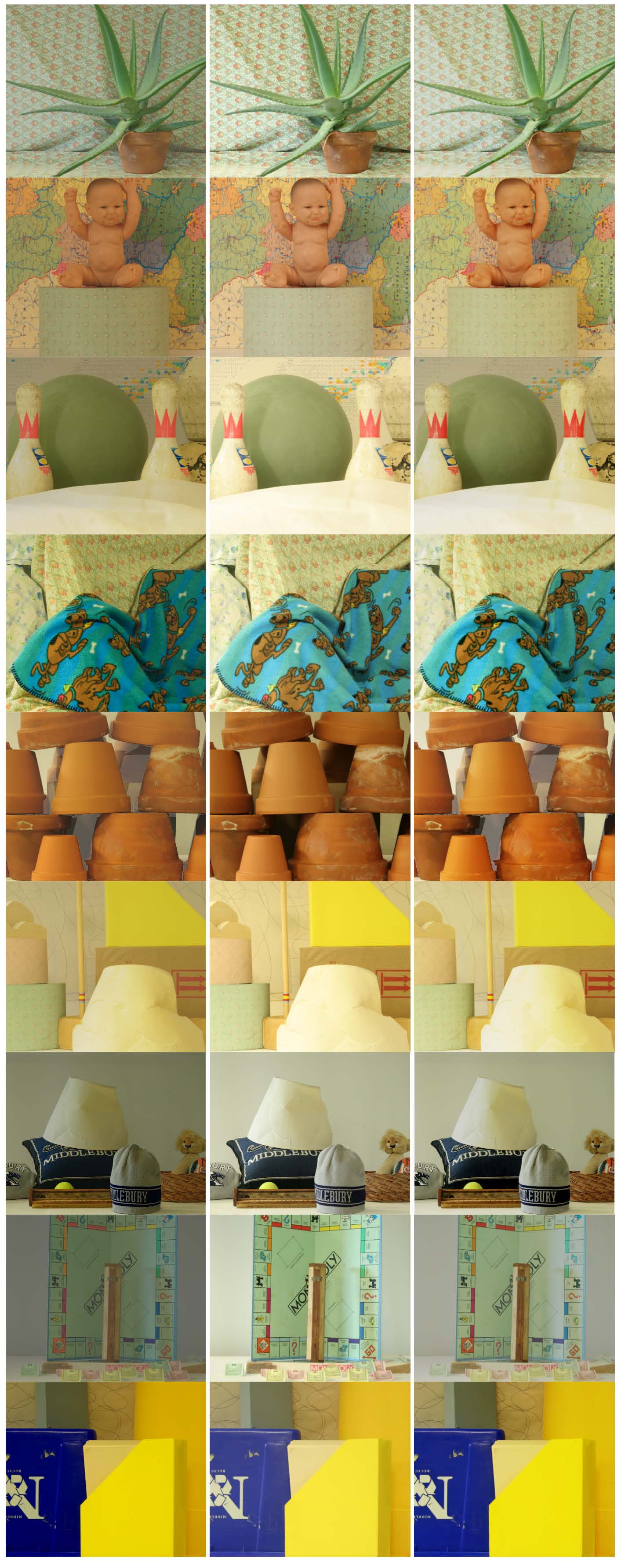}
        \caption{Dehazing results on ImageSet B (full-size comparison can be found in the supplementary material). Left: AOD results~\cite{li2017aod}. Middle: original images. Right: outputs of POGAN.}
        \label{fig:imagesetB}
\end{figure}

More importantly, we note that objective measures such as PSNR and SSIM often do not faithfully reflect the subjective quality of an image \cite{mittal2012no}. A convincing counterexample was reported in \cite{ledig2016photo} where the PSNR value of a perceptually much better image is dramatically (over $2dB$) lower than the benchmark; we note that this is also the reason for introducing adversarial loss in Eq. \eqref{equ:loss}. As shown in Figure \ref{fig:imagesetA} and \ref{fig:imagesetB}, it can be observed that our approach generates dehazing results with less color distortion and visually look more favorably than DCP/AOD (closer to the ground-truth - the middle column). 

\newcommand{\rowgroup}[2]{\hspace{-0.7em}#1\vspace{0.2em}#2}
\begin{table*}
\caption{Comparison of PSNR and SSIM values. Bold denotes the best in it's corresponding row.}
\label{tab:sync_AB}
\begin{tabular}{l l | a  b  a  b  a  b  a  b  a  b  a  b  a  b  a  b}
\hline
\rowcolor{LightCyan}
\mc{2}{Parameters}  & \mc{2}{DCP~\cite{he2011single}} & \mc{2}{NLD~\cite{berman2016non}} & \mc{2}{BCCR~\cite{meng2013efficient}} & \mc{2}{MSCNN~\cite{ren2016single}} & \mc{2}{DehazeNet~\cite{cai2016dehazenet}} & \mc{2}{AOD~\cite{li2017aod}} & \mc{2}{Ours}\\
\hline
\rowcolor{Cyan}
ImageSet A\\
\hline
\multirow{4}{*}{$\beta=0.75,~A=$}
& 0.7 & 	21.73 & 0.895 & 20.24 & 0.871 & 18.01 & 0.832 & 20.21 & 0.874 & 17.51 & 0.751 & 17.07 & 0.892 & 21.55 & 0.893\\

& 0.8 & 22.26 & 0.892 & 19.99 & 0.865 & 18.64 & 0.829 & 21.36 & 0.872 & 15.99 & 0.711 & 19.84 & 0.923 & 21.79 & 0.912\\

& 0.9 & 22.35 & 0.883 & 20.19 & 0.869 & 18.94 & 0.821 & 21.06 & 0.863 & 14.42 & 0.655 & 23.11 & 0.936 & 21.63 & 0.893\\

& 1   & 22.04 & 0.874 & 19.18 & 0.858 & 18.95 & 0.812 & 19.57 & 0.851 & 13.05 & 0.594 & 22.91 & 0.932 & 20.96 & 0.879\\
\hline
\mc{2}{Average} & \textbf{22.09} & 0.886 & 19.91 & 0.865 & 18.63 & 0.823 & 20.55 & 0.865 &	15.24 &	0.677 &	20.73 &	\textbf{0.921} & 21.48 & 0.894\\
\hline
\multirow{4}{*}{$\beta=1,~A=$}
& 0.7 & 	21.47 & 0.884 & 19.96 & 0.866 & 17.89 & 0.817 & 20.46 & 0.856 & 16.51 & 0.701 & 17.11 & 0.865 & 20.61 & 0.909\\

& 0.8 & 22.03 & 0.878 & 20.62 & 0.871 & 18.58 & 0.815 & 20.66 & 0.851 & 14.88 & 0.653 & 19.88 & 0.891 & 20.13 & 0.911\\

& 0.9 & 22.13 & 0.871 & 19.85 & 0.868 & 18.77 & 0.806 & 19.23 & 0.836 & 13.16 & 0.582 & 20.98 & 0.893 & 21.22 & 0.891\\

& 1   & 21.28 & 0.861 & 18.76 & 0.846 & 18.52 & 0.791 & 17.03 & 0.813 & 11.62 & 0.513 & 18.31 & 0.877 & 20.96 & 0.886\\
\hline
\mc{2}{Average} & \textbf{21.72} & 0.873 & 19.79 & 0.862 & 18.44 & 0.807 & 19.34 & 0.839 &	14.04 & 0.612 & 19.07 &	0.881 & 20.73 & \textbf{0.899}\\
\hline
\multirow{4}{*}{$\beta=1.25,~A=$}
& 0.7 & 21.35 & 0.873 & 19.67 & 0.865 & 17.74 & 0.801 & 19.91 & 0.825 & 14.65 & 0.659 & 16.82 & 0.812 & 20.74 & 0.812\\

& 0.8 & 21.93 & 0.872 & 20.16 & 0.864 & 18.43 & 0.802 & 19.02 & 0.816 & 14.05 & 0.61 & 18.88 & 0.832 & 20.51 & 0.834\\

& 0.9 & 21.84 & 0.862 & 19.82 & 0.861 & 18.45 & 0.792 & 17.21 & 0.798 & 12.19 & 0.524 & 18.22 & 0.827 & 19.36 & 0.856\\

& 1   & 20.25 & 0.852 & 17.91 & 0.829 & 18.05 & 0.776 & 14.79 & 0.769 & 10.49 & 0.447 & 14.92 & 0.804 & 20.86 & 0.883\\
\hline
\mc{2}{Average} & \textbf{21.34} & \textbf{0.864} & 19.39 & 0.854 & 18.16 & 0.792 & 17.73 & 0.802 &	12.84 &	0.561 & 17.21 & 0.818 & 20.36 & 0.846\\
\hline
\multirow{4}{*}{$\beta=1.5,~A=$}
& 0.7 & 	21.13 & 0.863 & 19.34 & 0.853 & 17.67 & 0.786 & 18.64 & 0.788 & 14.92 & 0.621 & 16.35  & 0.748 & 19.44 & 0.822\\

& 0.8 & 21.78 & 0.866 & 20.09 & 0.869 & 18.24 & 0.784 & 17.44 & 0.771 & 13.45 & 0.571 & 17.55 & 0.761 & 18.36 & 0.815\\

& 0.9 & 21.18 & 0.855 & 19.02 & 0.842 & 18.11 & 0.779 & 15.41 & 0.751 & 11.44 & 0.482 & 15.93 & 0.751 & 18.27 & 0.804\\

& 1   & 18.89 & 0.842 & 17.52 & 0.838 & 17.55 & 0.766 & 13.04 & 0.718 & 9.91 & 0.409 & 12.61 & 0.726 & 17.91 & 0.781\\
\hline
\mc{2}{Average} & \textbf{20.74} & \textbf{0.856} & 18.99 & 0.851 & 17.89 & 0.778 & 16.13 & 0.757 & 12.43 & 0.521 & 15.61 & 0.746 & 18.49 & 0.805\\
\hline
\rowcolor{Magenta}
ImageSet B\\
\hline
\multirow{4}{*}{$\beta=0.75,~A=$}
& 0.7 & 14.83 & 	0.791 & 15.39 & 0.718 & 13.47 & 0.724 & 16.61 & 0.836 & 22.39 & 0.887 & 15.80 & 0.925 & 23.32 & 0.883\\

& 0.8 & 15.82 & 0.807 & 16.58 & 0.738 & 14.29 & 0.744 & 18.11 & 0.858 & 21.36 & 0.873 & 18.23 & 0.949 & 22.18 & 0.922\\

& 0.9 & 16.99 & 0.841 & 17.01 & 0.771 & 15.16 & 0.769 & 19.59 & 0.881 & 20.05 & 0.856 & 21.53 & 0.963 & 21.65 & 0.898\\

& 1   & 17.84 & 0.864 & 17.81 & 0.761 & 15.92 & 0.779 & 20.58 & 0.891 & 18.54 & 0.831 & 25.03 & 0.967 & 20.16 & 0.884\\
\hline
\mc{2}{Average} & 16.37	& 0.825 & 16.69 & 0.747 & 14.71 & 0.754 & 18.72 & 0.866 & 20.58 & 0.861 & 20.14 &	\textbf{0.951} & \textbf{21.82} & 0.896\\
\hline
\multirow{4}{*}{$\beta=1,~A=$}
& 0.7 & 	14.49 & 0.748 & 15.44 & 0.685 & 13.17 & 0.681 & 16.82 & 0.807 & 22.17 & 0.874 & 15.76 & 0.921 & 21.92 & 0.892\\

& 0.8 & 15.82 & 0.785 & 16.01 & 0.673 & 14.12 & 0.705 & 18.75 & 0.835 & 20.91 & 0.863 & 18.91 & 0.948 & 20.02 & 0.901\\

& 0.9 & 17.25 & 0.844 & 16.58 & 0.703 & 15.26 & 0.754 & 20.39 & 0.867 & 19.19 & 0.836 & 22.93 & 0.961 & 18.99 & 0.865\\

& 1   & 18.31 & 0.864 & 17.52 & 0.752 & 16.16 & 0.774 & 20.61 & 0.881 & 16.91 & 0.793 & 23.19 & 0.961 & 19.14 & 0.873\\
\hline
\mc{2}{Average} & 16.46 & 0.811 & 16.38 & 0.703 & 14.67 & 0.728 & 19.14 & 0.847 &	19.79 & 0.841 &	\textbf{20.19} &	\textbf{0.947} & 20.01 & 0.882\\
\hline
\multirow{4}{*}{$\beta=1.25,~A=$}
& 0.7 & 	14.21 & 0.702 & 15.17 & 0.608 & 12.91 & 0.631 & 16.97 & 0.766 & 21.94 & 0.859 & 15.61 & 0.899 & 19.06 & 0.878\\

& 0.8 & 15.91 & 0.771 & 16.75 & 0.662 & 13.98 & 0.665 & 19.17 & 0.799 & 20.54 & 0.853 & 19.01 & 0.929 & 17.44 & 0.845\\

& 0.9 & 17.63 & 0.842 & 17.17 & 0.667 & 15.28 & 0.731 & 20.63 & 0.843 & 18.36 & 0.814 & 22.66 & 0.941 & 18.87 & 0.816\\

& 1   & 18.74 & 0.862 & 18.10 & 0.767 & 16.28 & 0.768 & 19.57 & 0.859 & 15.91 & 0.755 & 20.09 & 0.935 & 17.69 & 0.833\\
\hline
\mc{2}{Average} & 16.62 & 0.794 & 16.79 & 0.676 & 14.61 & 0.698 & 19.08 & 0.816 & 19.18 & 0.821 & \textbf{19.34} &	\textbf{0.926} & 18.26 & 0.843\\
\hline
\multirow{4}{*}{$\beta=1.5,~A=$}
& 0.7 & 	14.04 & 0.662 & 15.01 & 0.607 & 12.72 & 0.583 & 16.98 & 0.718 & 21.61 & 0.841 & 15.41 & 0.866 & 20.04 & 0.822\\

& 0.8 & 15.94 & 0.747 & 15.94 & 0.618 & 13.87 & 0.619 & 19.27 & 0.753 & 20.19 & 0.838 & 18.91 & 0.897 & 18.19 & 0.814\\

& 0.9 & 18.04 & 0.836 & 16.91 & 0.663 & 15.36 & 0.705 & 20.19 & 0.812 & 17.75 & 0.799 & 21.27 & 0.906 & 16.95 & 0.791\\

& 1   & 18.91 & 0.853 & 18.35 & 0.745 & 16.43 & 0.771 & 18.13 & 0.837 & 14.73 & 0.714 & 17.61 & 0.897 & 17.71 & 0.805\\
\hline
\mc{2}{Average} & 16.73 & 0.774 & 16.55 & 0.658 & 14.59 & 0.669 & \textbf{18.64} & 0.78 & 18.57 & 0.798 & 18.31 & \textbf{0.891} & 18.22 & 0.808 \\
\hline
\end{tabular}
\end{table*}

\subsection{Comparison against State-of-the-Art on Real-World Images}

\begin{figure*}[t]
        \includegraphics[width=\linewidth]{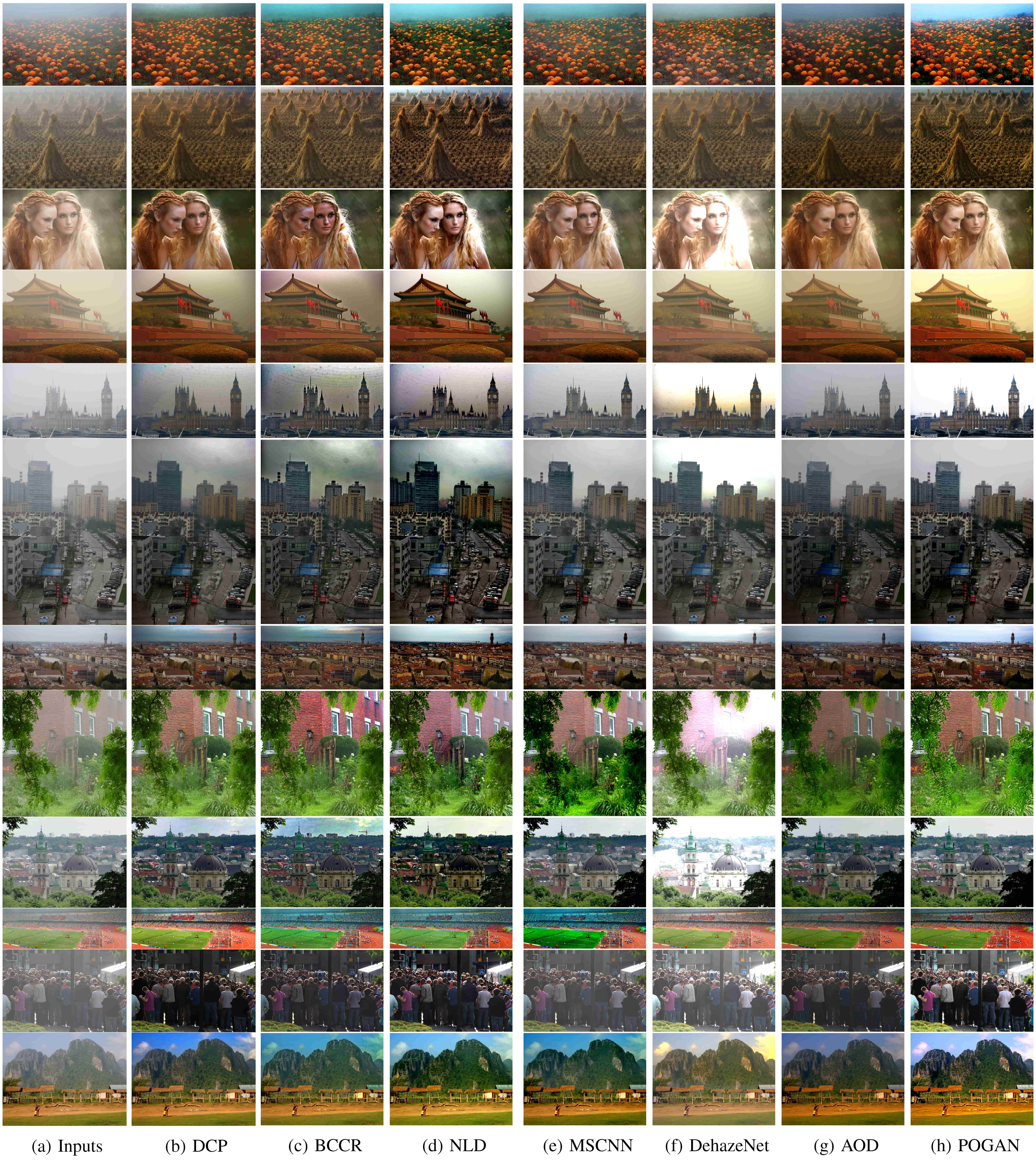} 
        \caption{Dehazing results on real-world images (full-size comparison can be found in the supplementary material).}
        \label{fig:realImg}
\end{figure*}

We have also compared our method against six state-of-the-art dehazing approaches on real-world hazy images (ImageSet C) as shown in Figure~\ref{fig:realImg}. This set of images contains large variations of scene depths and haze thickness as well as diverse scene structures such as portrait, landscape and architecture. Most of the images in ImageSet C have been included for evaluation purpose in previous studies of single image dehazing. We summarize the superiority of the proposed POGAN-based dehazing as follows: 1) it is effective at removing heavy haze in the presence of large scene depth variations such as the second row in Figure~\ref{fig:realImg}; 2) it significantly outperform other competing methods in terms of restoring color fidelity and vividness such as the second to the last  row in Figure~\ref{fig:realImg}\footnote{A more comprehensive comparison between this work and other competing approaches at the original resolutions has been included into the supplementary material accompanying this paper.}.

\section{Conclusions}
In this paper, we have presented a novel perceptually optimized GAN-based approach toward single image dehazing. Our approach directly learns a nonlinear mapping from the space of hazy images to that of haze-free ones using a deep residue network without estimating transmission maps. By casting the haze-free image as the fixed-point, we can recursively update the residue estimate until the convergence. To ensure visual quality, a discriminative network is introduced for adversarial learning and an adaptive perceptual loss function is developed to handle varying hazy conditions. Moreover, we proposed a novel application of guided filtering into the suppression of halo-like artifacts in dehazed images. Our extensive experimental results have shown that the subjective qualities of dehazed images by our perceptually optimized GAN (POGAN) are often more favorable than those by existing state-of-the-art approaches. The PSNR/SSIM performances of POGAN are also highly competitive especially when the hazy condition varies.

\bibliographystyle{IEEEtran}
\bibliography{IEEEabrv,mybib}

\end{document}